\documentclass[10pt,twocolumn,letterpaper]{article}

\usepackage{iccv}      % To produce the REVIEW version
\usepackage{multirow} 
\usepackage{titletoc}
\usepackage{amsmath}
\usepackage{bm}

\makeatletter
\DeclareRobustCommand\onedot{\futurelet\@let@token\@onedot}
\def\@onedot{\ifx\@let@token.\else.\null\fi\xspace}

\def\ie{\emph{i.e}\onedot}

\makeatother
\definecolor{iccvblue}{rgb}{0.21,0.49,0.74}

\usepackage{xcolor}
\usepackage[T1]{fontenc}
\definecolor{my_green}{RGB}{51,102,0}
\definecolor{my_red}{RGB}{204, 0, 0}
\definecolor{paired-light-blue}{RGB}{198, 219, 239}
\definecolor{paired-dark-blue}{RGB}{49, 130, 188}
\definecolor{paired-light-orange}{RGB}{251, 208, 162}
\definecolor{paired-dark-orange}{RGB}{230, 85, 12}
\definecolor{paired-light-green}{RGB}{199, 233, 193}
\definecolor{paired-dark-green}{RGB}{49, 163, 83}
\definecolor{paired-light-purple}{RGB}{218, 218, 235}
\definecolor{paired-dark-purple}{RGB}{117, 107, 176}
\definecolor{paired-light-gray}{RGB}{217, 217, 217}
\definecolor{paired-dark-gray}{RGB}{99, 99, 99}
\definecolor{paired-light-pink}{RGB}{222, 158, 214}
\definecolor{paired-dark-pink}{RGB}{123, 65, 115}
\definecolor{paired-light-red}{RGB}{231, 150, 156}
\definecolor{paired-dark-red}{RGB}{131, 60, 56}
\definecolor{paired-light-yellow}{RGB}{231, 204, 149}
\definecolor{paired-dark-yellow}{RGB}{141, 109, 49}  
\definecolor{myblue}{RGB}{218,232,252}
\definecolor{mygray}{RGB}{220,220,220}
\definecolor{mypink}{RGB}{251,49,153}

\usepackage{colortbl}
\usepackage{enumitem}
\usepackage{float}

\usepackage[pagebackref,breaklinks,colorlinks,allcolors=iccvblue]{hyperref}

\def\paperID{3662} % *** Enter the Paper ID here
\def\confName{ICCV}
\def\confYear{2025}

\title{Fast Autoregressive Video Generation with Diagonal Decoding}

\author{
\textbf{Yang Ye$^{*}$, Junliang Guo\thanks{~Equal contribution. Correspondence to Junliang Guo <junliangguo@microsoft.com>.}~, Haoyu Wu$^{*}$, Tianyu He, Tim Pearce, Tabish Rashid} \\ 
\textbf{Katja Hofmann, Jiang Bian} \\
Microsoft Research \\
~\\
\textbf{\normalsize{\url{aka.ms/diagd}}}
}

\begin{document}
\maketitle
\begin{abstract}
Autoregressive Transformer models have demonstrated impressive performance in video generation, but their sequential token-by-token decoding process poses a major bottleneck, particularly for long videos represented by tens of thousands of tokens. In this paper, we propose Diagonal Decoding~(DiagD), a training-free inference acceleration algorithm for autoregressively pre-trained models that exploits spatial and temporal correlations in videos. Our method generates tokens along diagonal paths in the spatial-temporal token grid, enabling parallel decoding within each frame as well as partially overlapping across consecutive frames. The proposed algorithm is versatile and adaptive to various generative models and tasks, while providing flexible control over the trade-off between inference speed and visual quality. Furthermore, we propose a cost-effective finetuning strategy that aligns the attention patterns of the model with our decoding order, further mitigating the training-inference gap on small-scale models. Experiments on multiple autoregressive video generation models and datasets demonstrate that DiagD achieves up to $10\times$ speedup compared to naive sequential decoding, while maintaining comparable visual fidelity.
\end{abstract}
    
\section{Introduction}
\label{sec:intro}
Recent advances in video generation models have achieved a significant level of performance in both diffusion~\citep{lin2024open, yang2024cogvideox, xu2024easyanimate} and autoregressive~\citep{kondratyuk2023videopoet, agarwal2025cosmos, kanervisto2025world} based methods. These models demonstrate impressive capabilities in learning foundational knowledge from raw videos and generating high-fidelity, controllable video outputs~\citep{videoworldsimulators2024}.
Consequently, video generation models have also been adopted in various domains in AI such as world modeling~\citep{ha2018world,agarwal2025cosmos,kanervisto2025world} and embodied AI~\citep{yang2023learning}, illustrating their potential power to serve as digital twins of the real world.

\begin{figure}[tbp]
   \centering
   \includegraphics[width=0.9\linewidth]{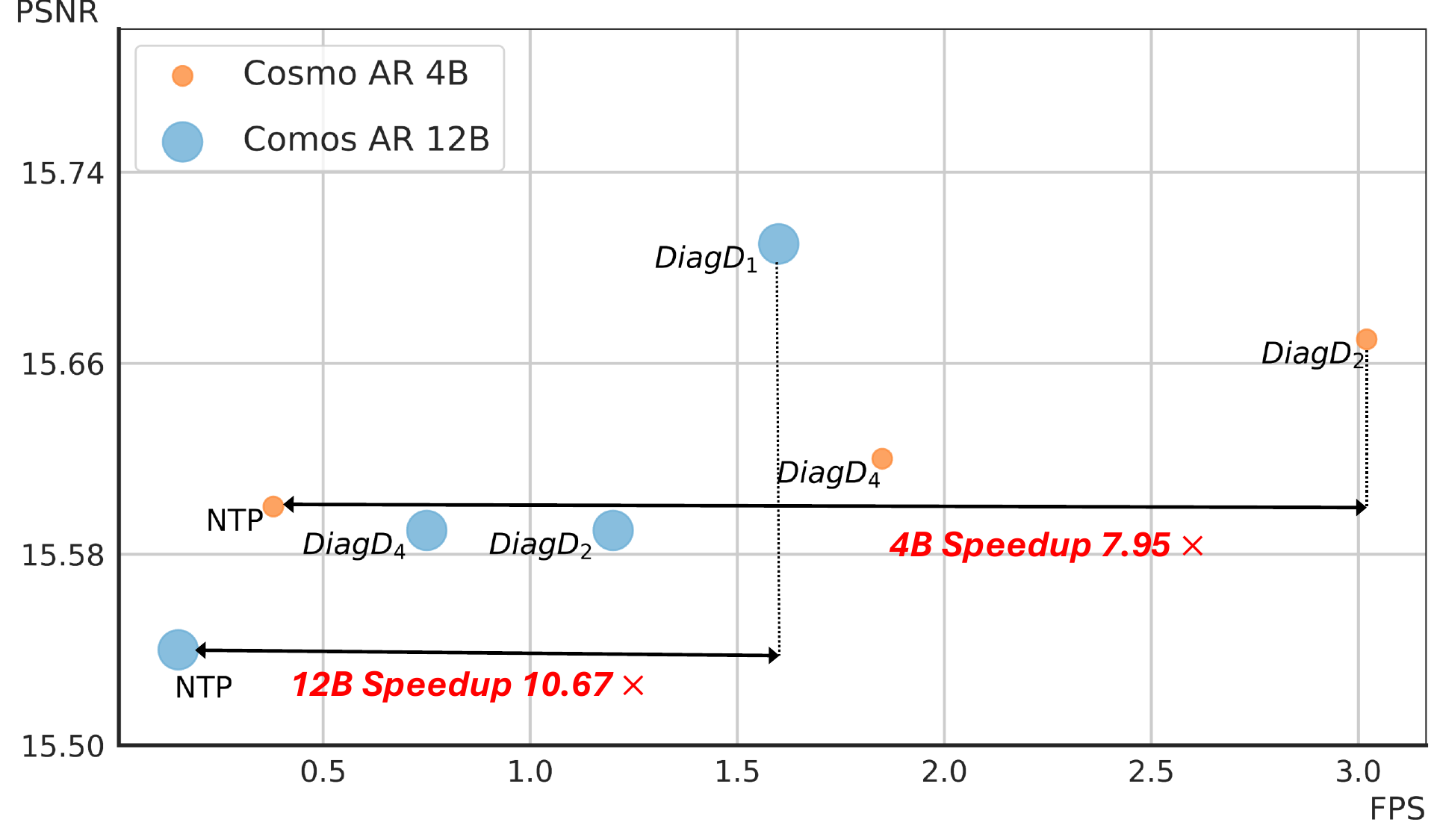}
   \vspace{-10pt}
   \caption{Comparisons between naive Next-Token Prediction~(NTP) and Diagonal Decoding~(DiagD) on Cosmos~\citep{agarwal2025cosmos} autoregressive models. The subscription of DiagD indicates different choices of the hyperparameters $k$. As a result, DiagD achieves a $10\times$ speedup with little degradation on visual quality.}
   \label{fig:headfig}
\end{figure}

Compared with diffusion models, autoregressive Transformers exhibit unique features as shown in the blooming of Large Language Models~(LLMs)~\citep{radford2019gpt2,brown2020language-gpt3} in recent years, including zero-shot emergent in-context learning capabilities~\citep{zhang2025autoregressive}, and scaling laws~\citep{kaplan2020scaling,pearce2024scaling}. Leveraging architectures similar to LLMs enables vision models to inherit these advancements and naturally extend to multi-modal inputs. Additionally, autoregressive models can generate videos of arbitrary length in a streaming paradigm, which is challenging for most of the diffusion models.

However, the video generation models usually utilize a visual tokenizer~\citep{esser2021taming, tang2024vidtok} to transform raw videos to tens of thousands of tokens, which poses a significant bottleneck for autoregressive models that generate tokens sequentially, especially when generating high-resolution, long-duration videos. The bottleneck can be divided into three main challenges.
Firstly, the naive next-token prediction mechanism leaves computational resources underutilized and thus leads to slow and costly generation. 
Secondly, previous visual autoregressive models~\citep{yu2023language, kondratyuk2023videopoet,bai2024sequential} generate tokens following a fixed raster-scan order (\ie, left-to-right, top-to-bottom, frame-by-frame), 
which creates suboptimal generation trajectories for image and video synthesis. 
Thirdly, the paradigm of autoregressive video generation model remains an underexplored area.

In this paper, we propose \textbf{Diagonal Decoding~(DiagD)}, an algorithm that utilizes redundant information in video representations by generating diagonal tokens in both spatial and temporal adjacent regions simultaneously. As illustrated in~\Cref{fig:diagonal}, instead of generating tokens sequentially in a raster-scan order, our method generates the diagonals of images from top left to bottom right, with tokens along the same diagonal produced in parallel at each step. By stacking frames together, diagonal decoding can be seamlessly applied to video generation. 
Notably, our method is training-free and functions as a plug-and-play module on autoregressively pretrained models, requiring only $5\%$ of the generation steps and up to a $10\times$ speedup in inference latency with negligible quality degradation, compared with next token prediction. We introduce hyperparameters to control the acceleration ratio in both spatial and temporal dimensions, allowing flexible adjustments to the trade-off between speed and performance.
Additionally, we also provide a finetuning strategy to enhance the generation quality especially for small scale models. 

We evaluate the performance and generalizability of Diagonal Decoding across various autoregressive video generation models, tasks, and datasets.  
Specifically, as shown in Figure~\ref{fig:headfig}, on the Cosmos~\citep{agarwal2025cosmos} world model, our method accelerates the inference by $10+$ times while achieving similar visual quality to next token prediction on tasks including video continuation and text-guided video generation.
On WHAM~\citep{kanervisto2025world}, a world model for games that produces multi-modal outputs, our spatial-only acceleration variant achieves approximately $4\times$ speedups while preserving generation quality. In addition, we also train autoregressive Transformer models from scratch to validate the performance of our method on different scales of models.

In summary, our contributions are three folds:
\begin{itemize}
    \item We propose \textbf{Diagonal decoding}, a plug-and-play acceleration algorithm for autoregressive video generation, achieving up to $10\times$ speedup in inference while maintaining generation quality.
    \item The proposed decoding algorithm demonstrates strong generalization capability across various autoregressive implementations including arbitrary visual tokenizers~(with or without temporal compression), arbitrary resolutions, and diverse generation tasks~(e.g. text-to-video, video continuation). 
    \item Besides inference, fine-tuning with diagonal decoding consistently improves the model performance, which provides inspiration for training video generation models in the future work.
\end{itemize}
\section{Related Work}
\label{sec:related}

\paragraph{Video Generative Models}
Video generative models have advanced rapidly in recent years, achieving impressive results in producing long-range, high-fidelity and controllable videos~\citep{ho2022imagen,kondratyuk2023videopoet,yu2023language-magvitv2,videoworldsimulators2024,liu2024videodpo,ma2024followyourpose,ma2024followyourclick}. These models have become powerful tools for related fields such as world models and embodied AI, enabling applications like game~\citep{kanervisto2025world} and agent simulation~\citep{yang2023learning}, and visual planning for robotics~\citep{du2023learning,chen2024igor}. 
Most video generation models consist of two components: a visual tokenizer~\citep{li2024wf,tang2024vidtok,wang2024vidtwin} that converts raw images and videos into latent representations, and a generative model that synthesizes these latents. However, representing a video clip requires a large number of latents. For instance, a $16$-frame video can produce between $40$k and $160$k tokens. The large number of latents creates a significant computational bottleneck for the generative model.

For the paradigm, diffusion~\citep{ho2020denoising,nichol2021improved} and autoregressive generation~\citep{vaswani2017attention} are the two most popular approaches in video generation. In this paper, we focus on autoregressive Transformers, as they have demonstrated performance on par with diffusion models~\citep{kondratyuk2023videopoet,yu2023language-magvitv2}, while inheriting key advantages from large language models, such as zero-shot in-context learning~\citep{zhang2025autoregressive}, long-range generation capabilities~\citep{liu2024world-lwm}, and the ability to smoothly integrate multiple modalities~\citep{kondratyuk2023videopoet}.

\paragraph{Parallel Decoding in Generative Models}
Parallel decoding has been widely explored in Transformer-based models to accelerate inference. Inspired by masked language models~\citep{ghazvininejad2019MaskPredict,guo-etal-2020-jointly}, MaskGIT~\citep{chang2022magit} and MAGVIT~\citep{yu2023magvit} introduce masked generative Transformers that generate tokens in parallel through iterative denoising.
Lformer~\citep{li2023lformer} divides tokens into several L-shaped blocks in each image, and generate tokens in each block in parallel. However, this method requires training the model from scratch. More recently, ZipAR~\citep{he2024zipar} proposes a parallel decoding algorithm for image generation by exploiting local token dependencies. Different from previous works, our diagonal decoding method is training free, operates at the video level, and achieves greater speedup ratios by handling temporal dependencies directly. \looseness=-1
\section{Method}
\label{sec:method}

\begin{figure*}[tbp]
  \centering
   \includegraphics[width=0.9\linewidth]{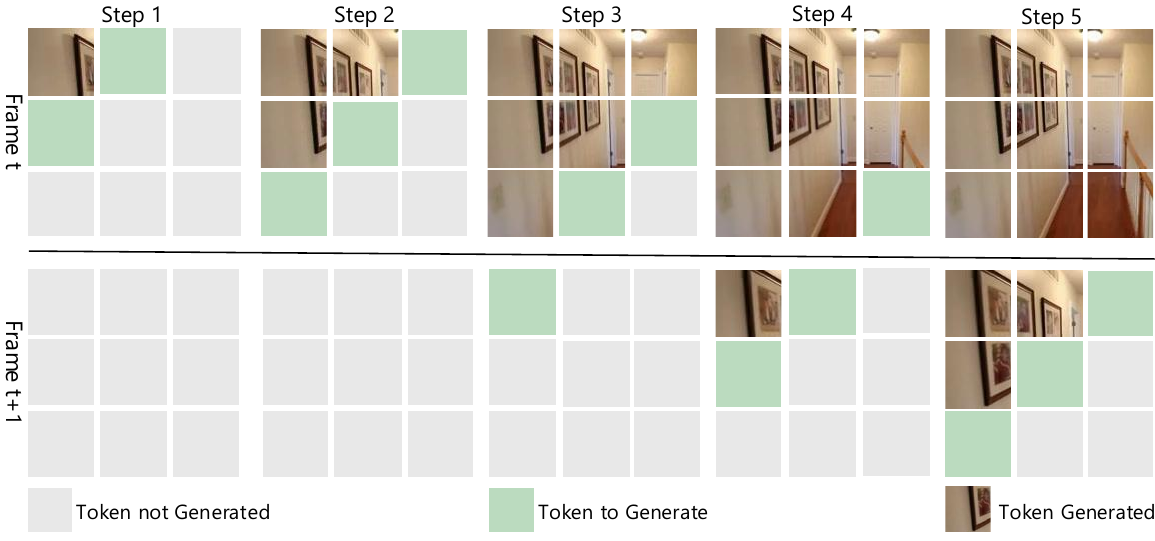}
   \caption{An illustration of the proposed Diagonal Decoding algorithm with $d=3$ and $k=1$. Spatially, tokens along the same diagonal within each frame are generated in parallel. Temporally, our method generates the top-left tokens of the subsequent frame before completing the current frame.}
\label{fig:diagonal}
\end{figure*}

\subsection{Background} 
In this section, we introduce the proposed Diagonal Decoding method. We begin with an overview of the task. Given a raw video $\mathbf{x}$ composed of a sequence of frames, a discrete VAE is used to encode the frames into a sequence of discrete tokens $\mathbf{c}$:
\begin{equation}
\begin{aligned}
    \mathbf{x} &= (x_1, \cdots,x_T), \\
    \mathbf{c} &= (c_1, \cdots, c_{n}, c_{n+1}, \cdots, c_{2n}, c_{2n+1}, \cdots c_N),
\end{aligned}
\end{equation}
where $T$ denotes the number of frames, $n$ denotes the number of tokens to represent each frame, and $N=T\cdot n$ denotes the total number of encoded tokens.
The autoregressive Transformer processes this sequence and learns to model the spatial and temporal dynamics of the video through next-token prediction. The training objective is to maximize the joint probability of each token, where the model predicts the current token based on all previously generated tokens:
\begin{equation}
\max _{\theta} \quad p_{\theta}(\mathbf{c})=\prod_{i=1}^{N} p_{\theta}\left(c_{i} \mid c_{1}, c_{2}, \cdots, c_{i-1}\right),
\label{equ:object}
\end{equation}
where $p_{\theta}$ denotes the Transformer model parameterized by $\theta$. 
During inference, the model generates tokens sequentially through next-token prediction, which is identical to a raster-scan order once the sequence is reshaped back into a $2$D structure. Finally, the decoder of the discrete VAE reconstructs the predicted tokens into videos in the RGB space.

\subsection{Diagonal Decoding}
\label{sec:method-diag}

The motivation of our method arises from intuitive observations on consecutive frames in a video, which can be summarized in two key insights. As shown in~\Cref{fig:diagonal}, the first insight is that patches exhibit stronger correlations with their spatial neighbors than with sequential ones. For example, the first patch in each row is more related to the first patch in the previous row than to the last patch in the same row, despite the latter being its sequential predecessor. Secondly, due to the temporal redundancy of videos, patches from consecutive frames that occupy similar relative positions are highly likely to be similar to each other. We empirically validate our observations in~\Cref{fig:attentionmap} in experiments.
As a result, we find that sequential autoregressive generation is not only counterintuitive but also inefficient, and we propose leveraging these spatial and temporal correlations to accelerate the generation process.

Specifically, we propose Diagonal Decoding, an iterative algorithm that generates tokens along diagonal paths in the spatial-temporal token grid. Spatially, within each frame, tokens along the same diagonal are generated in parallel, leveraging the strong local dependencies between neighboring patches.
And temporally, as illustrated in~\Cref{fig:diagonal}, by stacking frames together, our method generates the top-left tokens of the next frame before completing the current frame, as these tokens are less likely to depend on the bottom-right tokens that have not yet been generated.

Formally, let $h$ and $w$ denote the height and width of a frame, respectively, and let $c_{t,i,j}$ represent the token corresponding to the patch in row $i$ and column $j$ of the $t$-th frame in the video. We introduce two hyperparameters to define our algorithm. The parameter $k$ represents the number of existing spatial neighbors in the previous row available when generating the current token. In other words, $c_{t,i,j}$ is generated after all tokens satisfying $c_{t, \leq i, \leq j+k}$ have been generated. Then we can calculate the number of iterations to generate all tokens in one frame:
\begin{equation}
    s_{\textrm{spa}} = (h-1)\cdot k + w.
\end{equation}
Compared to the standard next-token prediction, which requires $h \cdot w$ steps, the acceleration ratio of spatial diagonal decoding is given by:
\begin{equation}
    r_{\textrm{spa}} = \frac{h\cdot w}{(h-1)\cdot k + w} \approx \frac{\min \{h,w\}}{k+1}. 
\label{equ:ratio_spatial}
\end{equation}

From the temporal aspect, we introduce temporal delay $d$ to represent the number of diagonal lines that must be generated in the previous frame before starting the next frame. The value of $d$ ranges in $[1, s_{\textrm{spa}}]$, where $d=1$ represents the extreme case where the next frame begins generating immediately after the first token of the previous frame is produced, and $d=s_{\textrm{spa}}$ corresponds to no temporal level acceleration being utilized. Equipped with both spatial and temporal diagonal decoding, the number of iterations to generate all tokens in $T$ frames can be written as:
\begin{equation}
    s_{\textrm{tep}} = (T-1)\cdot d + s_{\textrm{spa}}.
\end{equation}

As illustrated in~\Cref{fig:diagonal}, we set $k=1$ and $d=k\cdot h$ to balance generation quality and speed in most of our experiments, while naturally aligning spatial and temporal acceleration within a unified diagonal decoding framework. The total speedup ratio compared to next-token prediction is given by:
\begin{equation}
    r_{\textrm{diag}} = \frac{T\cdot h \cdot w}{(T-1)\cdot h + h+w-1} \approx w.
\label{equ:ratio_diag}
\end{equation}
The derivation of Equation~(\ref{equ:ratio_spatial}) and (\ref{equ:ratio_diag}) are shown in the appendix.
This indicates that the acceleration is roughly proportional to the width of the video resolution, significantly reducing the number of decoding iterations relative to standard autoregressive methods. We provide analysis on both hyperparameters $k$ and $d$ in experiments.

\paragraph{Discussion}
The two hyperparameters, $k$ and $d$, flexibly control the tradeoff between inference speed and generation quality, enhancing the versatility of our method. For pure video generation models, such as Cosmos~\citep{agarwal2025cosmos}, both spatial and temporal acceleration can be enabled to achieve the highest inference speed. 
On the other hand, for models with multimodal outputs~(e.g., WHAM~\citep{kanervisto2025world}, which generates paired images and actions in games) where temporal acceleration is not applicable, setting $d=s_{\textrm{spa}}$ allows the model to leverage spatial diagonal decoding alone. This adaptability makes our framework suitable for a wide range of generative models and tasks.

The spatial-only variant of our diagonal decoding shares insights with ZipAR~\citep{he2024zipar}, a decoding algorithm for text-to-image generation, but introduces a key innovation: leveraging temporal redundancies across frames to enhance efficiency even for the spatial-only algorithm. 
Specifically, implementing diagonal decoding introduces a training-inference gap, as the first token in each row is conditioned on the last generated token of the previous row during training, and such dependency is absent during inference. Previous methods use the last token in the previous row, $c_{t,i-1,j+k}$, as the predecessor for generating token $c_{t,i,j}$. In contrast, our approach leverages temporal information by using $c_{t-1,i,j}$, the token at the same position in the previous frame, which provides additional context and enables more accurate predictions. As a result, while ZipAR requires a large $k=16$ to maintain visual quality, our spatial-only diagonal decoding achieves higher speedups (e.g., $k=1$ on large scale models or $2$ on smaller ones) without compromising visual fidelity.

\subsection{Finetuning Strategy}
For small-scale models with limited capacity, the training-inference gap discussed above may lead to performance degradation. To mitigate this, we propose two solutions. First, as described in~\Cref{sec:method-diag}, the hyperparameters $k$ and $d$ can be tuned to balance visual quality and generation speed. As shown in the experiments, simply increasing $k$ from $1$ to $2$ significantly improves visual quality while preserving fast inference.

Second, we introduce a cost-effective finetuning strategy that replaces the standard causal attention mask with one aligned to our diagonal decoding algorithm. In experiments, we observe that finetuning for just $1$k steps significantly reduces performance degradation, making this approach both practical and efficient.

\section{Experiment}
\label{sec:exp}

\begin{table*}[tb]
\centering
    \caption{Quantitative evaluation on Cosmos. $4$B and $12$B refer to models used for video continuation, while $5$B and $13$B refer to models used for text-to-video generation. "NTP" refers to the next-token prediction paradigm. DiagD $\ k=i$ denotes the Diagonal Decoding algorithm where $d$ is equal to $k\cdot h$ and $k$ equals different values of $i$. "STEP" refers to the number of forward passes required by the model to generate a video. "TP" represents throughput, i.e., number of tokens the autoregressive model can generate per second.}
    {
    	\begin{tabular}{lcl|cccc|ccc}
    		\toprule[1pt]
    	   Tasks &Model& Algorithm  & FVD$\downarrow$ & LPIPS$\downarrow$ & SSIM$\uparrow$ & PSNR$\uparrow$  & FPS$\uparrow$  &TP$\uparrow$ & STEP~($k$)$\downarrow$\\
            \midrule[1pt]
    		Text-to-Video   &14B      & Diffusion       & \bm{$129$} & \bm{$0.43$} & 0.60 & 15.15  &0.08   & /   &  / \\
            \midrule[1pt]
    		\multirow{3}{*}{Continuation }   & \multirow{3}{*}{4B} & NTP    & 136 & 0.44 & 0.63& 15.60  & 0.38  &120  &$7.68$   \\
                                            &     & DiagD $k=2$           & 137 & 0.44  & 0.63 & 15.67  & 3.02  &966   &$0.30$   \\
                                            &    & DiagD $k=1$            & 348 & 0.62  & 0.61 & 13.27  & \bm{$4.00$}  &\bm{$1280$}  &\bm{$0.18$}     \\
            \midrule
    		\multirow{3}{*}{Continuation}   &\multirow{3}{*}{12B}& NTP                   & 135 & 0.44 & 0.63 & 15.54  &0.15   &49   &$7.68$  \\
                               & & DiagD $k=2$           & 136 & 0.44 & 0.63  & 15.59  &1.21   &384  &$0.30$  \\
                           &   & DiagD $k=1$      & 136 & 0.44  & 0.63 & \bm{$15.71$}  &1.62   &512  &\bm{$0.18$} \\
            \midrule[1pt]
    		\multirow{3}{*}{Text-to-Video}   &\multirow{3}{*}{5B}& NTP                    & 137 &0.44  & 0.63& 15.57  &0.34   &108  &$7.68$ \\
                               & & DiagD $k=2$           & 151 &0.45  & 0.62& 15.38  &2.67   &853  &$0.30$  \\
                               & & DiagD $k=1$           & 173 & 0.47 & 0.62& 15.16  &3.43   &1097 &\bm{$0.18$} \\
            \midrule
            \multirow{3}{*}{Text-to-Video}   &\multirow{3}{*}{13B}& NTP         & 137 & 0.44 & 0.63 & 15.56 &0.12   &39   &$7.68$ \\
                       & & DiagD $k=2$                   & 153 & 0.44 & 0.62& 15.41  &1.04   &334  &$0.30$  \\
                       & & DiagD  $k=1$                  & 158 & 0.44  &0.63 & 15.52  &1.20   &384  &\bm{$0.18$} \\
    
    		\bottomrule[1pt]
    	\end{tabular}
    }
    \label{tab:cosmos-result}
 \vspace{-2mm}
\end{table*}

In this section, we present experimental results on the proposed Diagonal Decoding algorithm. We start with experimental setups described in~\Cref{sec:exp-setup}, followed by the main results on various models and datasets illustrated in Section~\ref{subsec:main}. Finally, we provide analysis and case studies in~\Cref{subsec:analysis} and~\Cref{subsec:case}.

\subsection{Setups}
\label{sec:exp-setup}

\subsubsection{Baselines}
We consider models that produce both pure video and multi-modal outputs as our baselines to validate the performance of Diagonal Decoding with temporal and spatial accelerations respectively. To study the relations between model scales and the performance of DiagD, we also train autoregressive models from scratch.

\paragraph{Cosmos} Cosmos~\citep{agarwal2025cosmos} is a world foundation model collection that integrates multiple pre-trained models. We utilize the released autoregressive models, which are equipped with a discrete video tokenizer that provides $8\times$ temporal compression and $16\times$ on spatial. As a result, for $8$ frames with $640\times1024$ as the raw resolution, it is encoded into latent discrete tokens with size $40\times 64=2,560$, i.e., $h=40$ and $w=64$ following our notations. Experiments on Cosmos show the generalability of Diagonal Decoding on representations with temporal compressions.

\paragraph{WHAM} 
The World and Human Action Model (WHAM)~\citep{kanervisto2025world} is a recently proposed state-of-the-art autoregressive generative model on the game environment, which is capable of generating accurate and coherent game scenes following instructions from users. Different from Cosmos which produces videos solely, WHAM takes interleaved concatenations of images and actions as input and output, to receive controls and generate consequences. Therefore, WHAM utilizes an image tokenizer with $10\times$ spatial compression only, which transforms a raw game scene with $180\times 300$ into $18\times 30=540$ tokens, with $h=18$ and $w=30$ as a result.
We testify DiagD with spatial acceleration on WHAM, considering that the action will be given by the user only after the previous game scene has been generated.

\paragraph{MC-AR} 
To study the performance of Diagonal Decoding on different scales of models, as well as validating the proposed fine-tuning strategy, we train a series of models from scratch. Specifically, we utilize the VPT dataset~\citep{baker2022video} which consists of pairs of game scenes and actions on the game Minecraft. We transform the raw game scenes with an image VQ-VAE~\citep{patil2024amused} to latent tokens with size $14\times 24=336$, i.e., $h=14$ and $w=24$. Then, a Transformer decoder is trained with next token prediction by taking the concatenation of game scenes and actions as input. We train model scales from $300$M to $1.2$B parameters. We leave detailed descriptions of baselines and the training procedure in the appendix.

\subsubsection{Evaluation Setups} 

\begin{table}[tb]
    \centering
    \caption{Quantitative evaluation on WHAM. Each evaluation video has a duration of $10$ seconds and a frame rate of $10$ fps. For every video in this dataset, the initial ten frames along with the complete action sequence serve as prompts for generation.}
    \vspace{-2mm}
    \resizebox{\columnwidth}{!}
    {
    	\begin{tabular}{cl|cc|cc}
    		\toprule[1pt]
    	    WHAM & Algorithm  & PSNR$\uparrow$ & FVD$\downarrow$ & FPS$\uparrow$ & STEP~($k$)$\downarrow$ \\
            \midrule
    		\multirow{3}{*}{200M}            & NTP                             & \bm{$15.04$}      & \bm{$367$}   & $0.23$   & $54$    \\% 
                                & DiagD $k=2$                 & $14.08$      & $462$   &$0.75$   & $6.4$     \\
                                & DiagD $k=1$                 & $14.27$      &$472$   & \bm{$0.97$}   & \bm{$4.7$}     \\%  &   \\
            \midrule
    		\multirow{3}{*}{1.6B}           & NTP                              & \bm{$15.12$}      & \bm{$336$}   &$0.12$   & $54$     \\%  &  \\
                                & DiagD $k=2$                 & $14.92$      &$378$   & $0.34$   & $6.4$     \\
                               & DiagD  $k=1$                 & $14.77$      &$365$   & \bm{$0.40$}   & \bm{$4.7$}   \\% &  \\
            
    		\bottomrule[1pt]
    	\end{tabular}
    }
    \label{tab:wham-result}
 \vspace{-2mm}
\end{table}

\paragraph{Metrics}
For all models, we use one NVIDIA 80GB A100 GPU and batch size as $1$ to obtain results. We propose separate metrics to assess the visual quality and inference speed. We follow common practices and utilize metrics including Fréchet Video Distance (FVD)~\citep{unterthiner2018towards}, Peak Signal-to-Noise Ratio (PSNR)~\citep{hore2010image}, Learned Perceptual Image Patch Similarity (LPIPS)~\citep{zhang2018unreasonable}, and Structural Similarity Index Measure (SSIM)~\citep{wang2004image}. 
For the inference speed, we report three metrics. The Frames Per Second~(FPS) generated by the model, calulated with the wall-clock time. And Step that denotes the number of forward passes for a model to generate the video. In addition, ThroughPut~(TP) on output tokens per second is also recorded, which is a crucial metric in real-time applications.

In addition to automatic metrics, we also provide human evaluations, detailed in~\Cref{subsec:case}, to assess the generation results from various aspects including the general visual quality, object movement consistence, and the comparisons between two settings.

\paragraph{Evaluation Dataset}
The test set of each model is introduced below. For Cosmos, an open-sourced evaluation pipeline and datasets is absent. Therefore, we implement the pipeline by ourselves following details provided in their technical report~\citep{agarwal2025cosmos}. We randomly sample $100$ videos with $33$ frames from the RealEstate10K dataset~\citep{zhou2018stereo} as the test set. 
For WHAM, we randomly selected $100$ videos with $100$ frames from the official evaluation set due to the limitations on time and resource. Experiments on the whole test set with $1$k videos will be reported in future revisions.
As for MC-AR, we split $100$ video clips from VPT~\citep{baker2022video} as the test set, each containing $16$ frames. 
We leave more details in appendix. 

\subsection{Main Results}
\label{subsec:main}

\begin{table}[t]
    \caption{Quantitative evaluation on 700M MC-AR. Fine-tuning with DiagD attention mask helps bridge the training-inference gap and improve performance.
    }
    \label{tab:minecraft-result}
    \centering
    \vspace{-2mm}
    \resizebox{\columnwidth}{!}
    {
    	\begin{tabular}{l|cc|cc}
    		\toprule[1pt]
    	    Algorithm  & PSNR$\uparrow$ & FVD$\downarrow$ & FPS$\uparrow$ & STEP~($k$)$\downarrow$ \\
    		\midrule
    		NTP                               &\bm{$15.74$}       &\bm{$210$}   &1.08    &$5.04$    \\
            DiagD w/o FT                &15.27       &247   &\bm{$2.09$}    &\bm{$0.75$}        \\
            DiagD w/ FT        &15.32       &231   &\bm{$2.09$}    &\bm{$0.75$}      \\
    		\bottomrule[1pt]
    	\end{tabular}
    }
 \vspace{-2mm}
\end{table}

\begin{figure}[tbp]
  \centering
   \includegraphics[width=0.9\linewidth]{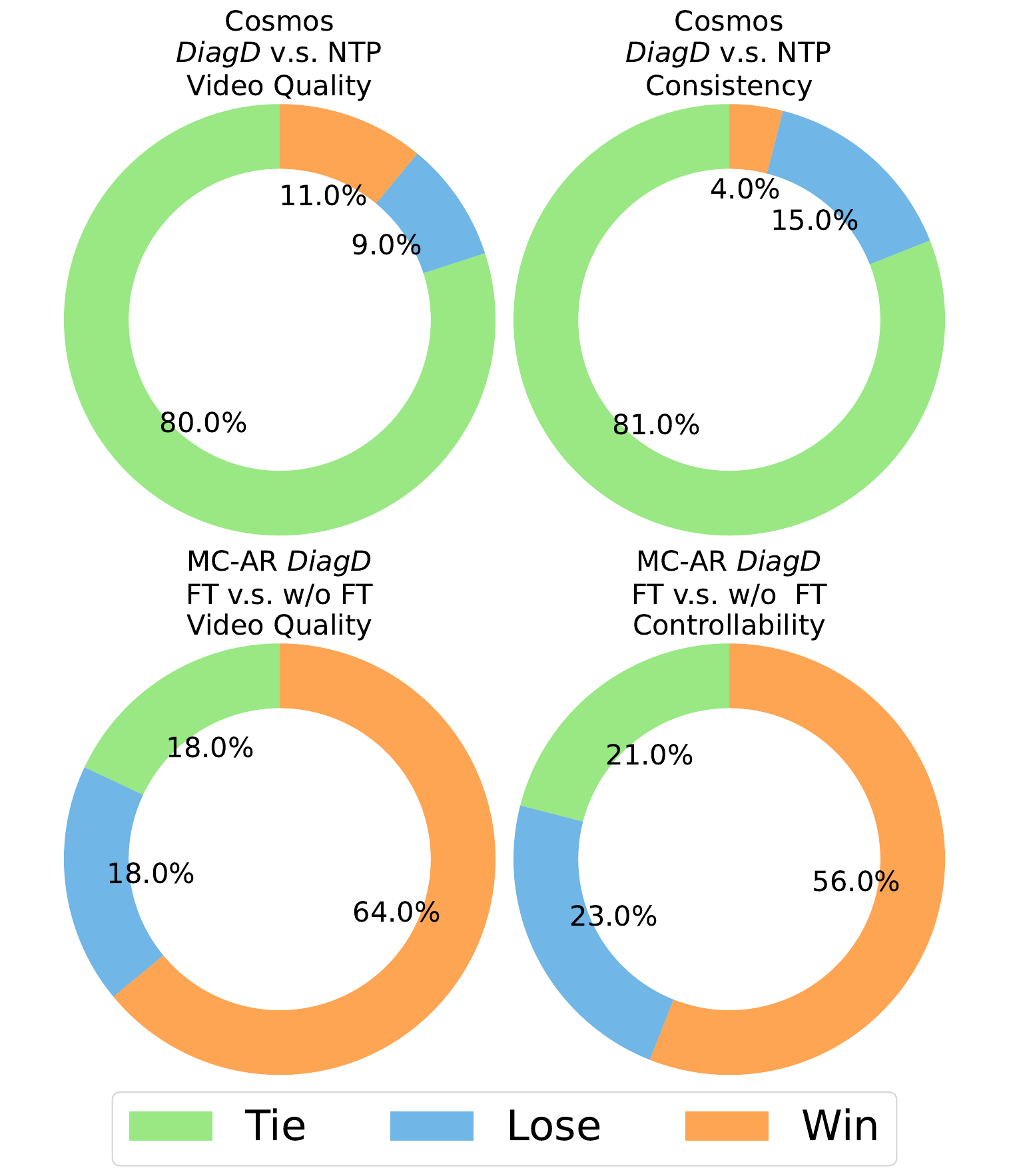}
   \caption{Human evaluation results for Cosmos-12B with DiagD and MC-AR 700M with or without DiagD finetuning. In the figure, "Win" indicates the left setting outperforms the right one, while "Lose" represents the opposite. The results indicate that DiagD achieves similar performance to NTP, and fine-tuning helps it perform even better.
}
   \label{fig:cosmos_human_eval_ring}
\end{figure}

\begin{figure}[tb]
  \centering
   \includegraphics[width=1\linewidth]{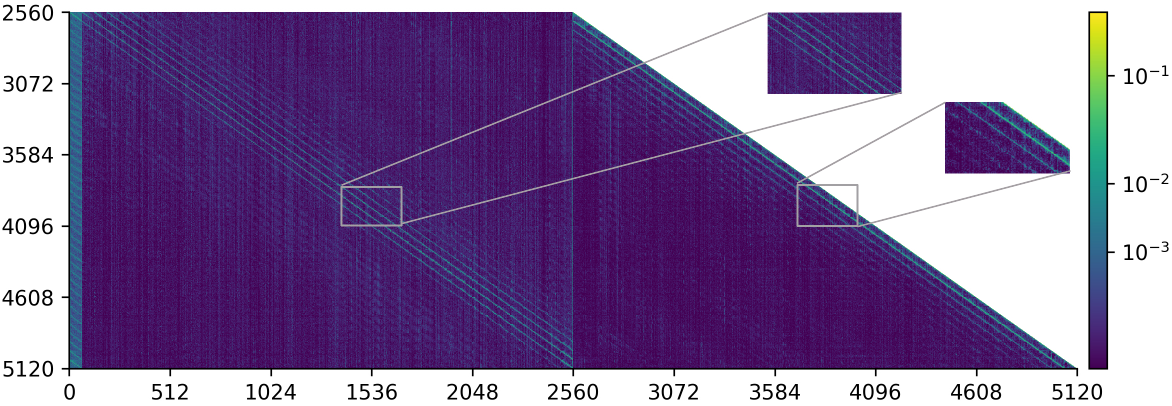}
   \caption{The attention scores of the second frame in the Cosmos-4B model are shown. The bright slash lines indicate that substantial attention scores are assigned to tokens at regular intervals, corresponding to those in temporally and spatially adjacent positions. The shown attention map is the mean value of all self-attention layers in the model.
}
   \label{fig:attentionmap}
\end{figure}

\begin{figure*}[!tb]
  \centering
   \includegraphics[width=0.9\linewidth]{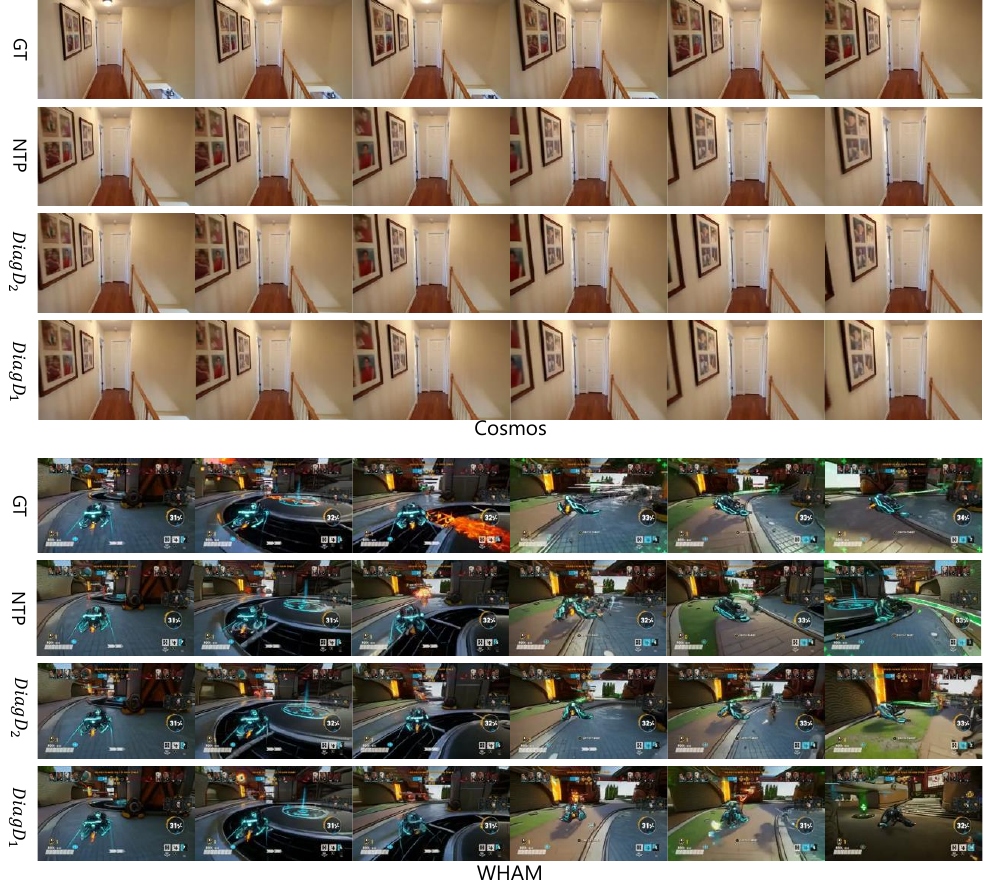}
   \caption{Qualitative analysis of Cosmos and WHAM. Videos generated by Cosmos-12B and 1.6B WHAM  models using the next-token prediction paradigm (second row) and Diagonal Decoding under different configurations (bottom two rows). The ground truth is presented in the first row. We sample every 6 frames from the generated videos in Cosmos and every 8 frames from those in WHAM.}
   \label{fig:maincase}
\end{figure*}

In this section, we illustrate the main results of DiagD on various baseline models and tasks. 
\paragraph{Cosmos} 
We apply DiagD with both temporal and spatial acceleration to Cosmos autoregressive models, setting $d = k \cdot h$ so that the next frame begins generating after the first $k$ tokens in the last row of the previous frame are produced. 
We consider two tasks including video continuation and text-to-video generation, as well as different scales of Cosmos models. Conditioned on $9$ initial frames~(and textual desciptions for the text-to-video task), the model is required to generate the following $24$ frames.
We also show the performance of the $14$B Cosmos Diffusion model for reference.

Results are listed in Table~\ref{tab:cosmos-result}, where the variants of DiagD with different $k$s are considered. The notation of $d$ is omitted as it is same across variants. Compared to naive Next-Token Prediction~(NTP), which requires $7.68$k steps to generate $24$ frames, DiagD reduces the step count to only $2\%$ to $4\%$, enabling substantial parallelism in decoding. In terms of FPS measured by wall-clock time, DiagD achieves approximately $10\times$ speedup over NTP across various settings and model scales.

For visual quality, the fastest variant~($k=1$) introduces degradation in the smaller 4B Cosmos model. However, this gap disappears either by setting $k=2$ or scaling up to the 12B model. Overall, Diagonal Decoding significantly accelerates inference speed in a training-free manner, with minimal impact on visual quality.

\paragraph{WHAM} 
On WHAM models, we validate the performance of DiagD with spatial acceleration solely. Specifically, we set $d=s_{spa}$ and $k$ to $1$ or $2$ in this setting. The generation task here is challenging as the model is required to generate $100$ frames conditioned on one initial frame and a sequence of actions, and slight errors in previous frames will cause huge performance drop due to error accumulation. 

The results are shown in Table~\ref{tab:wham-result}. The spatial only DiagD requires $10\%$ steps to generate the whole video compared to NTP, and brings around $4$ times speedups regarding FPS. This aligns with our derivations in Equations~(\ref{equ:ratio_spatial}) and (\ref{equ:ratio_diag}), where temporal acceleration introduces an additional speedup of roughly $k+1$ times. In terms of visual quality, the larger 1.6B model exhibits less performance degradation compared to the smaller 200M model, suggesting that larger models can better tolerate the training-inference gap introduced by diagonal decoding. Overall, these results demonstrate the effectiveness of the spatial variant of DiagD in balancing generation speed and visual fidelity across different model scales.

\paragraph{MC-AR}

We validate the proposed fine-tuning strategy on MC-AR models and analyze its scaling behavior in the appendix. Specifically, we replace the standard causal attention mask in the pre-trained autoregressive Transformer with one aligned to DiagD, then fine-tune the model for another $1$k steps. As shown in Table~\ref{tab:minecraft-result}, fine-tuning effectively mitigates the training-inference gap, enhancing generation quality while preserving the fast inference speed of DiagD.

\paragraph{Human Evaluation} 
We conduct human evaluations as a complementary to automatic evaluations. For Cosmos-12B, we provide $10$ videos generated by next-token prediction and DiagD, and ask participants to evaluate which one performed better in terms of the visual quality and camera consistency. For MC-AR, we ask participates to compare generation results from DiagD with or without fine-tuning, in terms of the visual quality and controllability. As shown in Figure~\ref{fig:cosmos_human_eval_ring}, we find that: 1) DiagD and NTP generate videos with similar visual quality and semantic meaning; 2) fine-tuning help improve the visual quality significantly for a smaller 700M model.

\begin{figure}[tbp]
   \centering
   \includegraphics[width=1\linewidth]{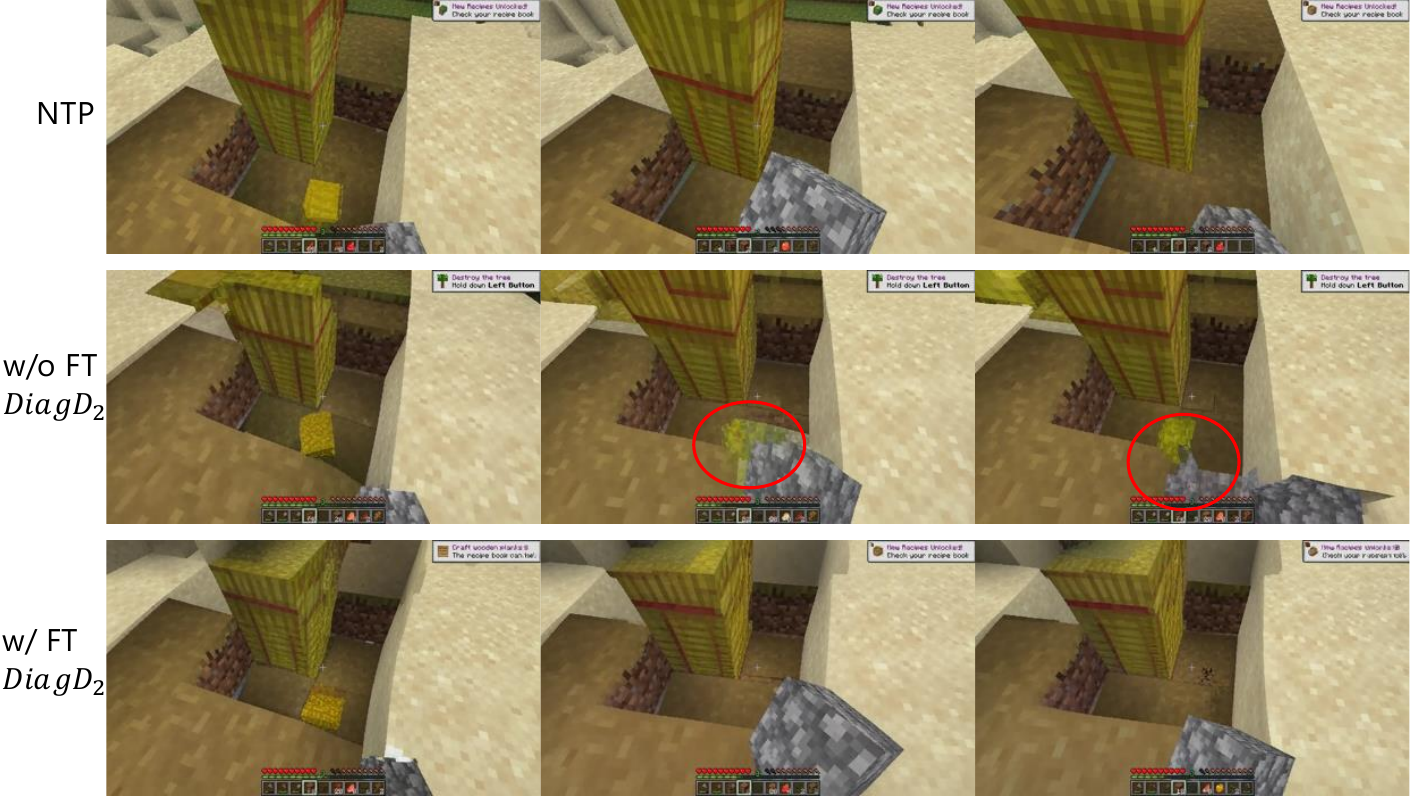}
   \caption{Qualitative comparison of results on MC-AR. Frames generated by models without fine-tuning may appear blurry, which can be mitigated with an additional $1k$-step fine-tuning.}
   \label{fig:finetune}
\end{figure}

\subsection{Analysis}
\label{subsec:analysis}
\paragraph{Attention Pattern} We visualize the attention map of the second frame generated by the Cosmos-Autoregressive-4B model, as depicted in~\Cref{fig:attentionmap}. The diagonal patterns indicate that significant attention scores are allocated to tokens at fixed intervals, corresponding to tokens located in the same column of previous rows and the preceding frame. Spatially, tokens along the same diagonal exhibit notably high attention scores, indicating strong spatial relevance, as shown in the right square highlighted in the attention map. Temporally, tokens primarily attend to adjacent positions in the previous frame, emphasizing temporal correlation, as illustrated by the left square in the attention map. The results of the attention map provide empirical support for the intuitive motivation introduced in Section~\ref{sec:method-diag}.

\paragraph{Scaling Effects} We also observe that DiagD achieves better performance and greater speedup on larger models, confirming that larger models capture more spatial and temporal properties in videos than smaller ones. Although Cosmos-4B and Cosmos-12B exhibit nearly identical FVD scores with next-token prediction, they demonstrate significantly different scores when using DiagD~($k = 1$). Additionally, we observed that both Cosmos-12B and WHAM-1.6B achieve higher FPS and superior visual quality with DiagD compared to their smaller counterparts employing next-token prediction. Therefore, DiagD may serve as an effective benchmark for evaluating whether models accurately capture spatial and temporal redundancies, which can reveal the nature of autoregressive video modeling. 

\paragraph{Study on Hyperparameters} In Table~\ref{tab:cosmos-result} in the Appendix, we present the results of various hyperparameter combinations for Cosmos-4B and Cosmos-12B. We found that $k$ has a more significant impact on controlling the speedup ratio than $d$. When $k$ values are similar, the FPS remains comparable. Additionally, increasing $d$ can enhance visual quality. By adjusting the values of $d$ and $k$, we can precisely balance visual quality and inference speed. This finding suggests that DiagD can be flexibly leveraged, eliminating the necessity of training smaller models solely for computational efficiency, decoupling the strong correlation between model size and inference speed.

\subsection{Case Study}
\label{subsec:case}

In this section, we present case studies on the generation results of DiagD on different models. In the top half of Figure~\ref{fig:maincase}, we show results using DiagD on the Cosmos-12B model. Compared to results obtained by next-token prediction, DiagD provides consistent camera movements and comprehensive image details. In the bottom half, we show the results of the spatial variant DiagD on the 1.6B WHAM model, where DiagD consistently provides high-fidelity images in all frames, without observing error accumulations. Furthermore, videos generated by next-token prediction and DiagD exhibit nearly identical object movements, indicating that our algorithm successfully preserves model controllability and visual quality during long-range generation.

In addition, we present cases to demonstrate the effectiveness of our fine-tuning method. As shown in Figure~\ref{fig:finetune}, fine-tuning for only $1$k steps helps alleviate the generation of blurry areas, and provides videos with similar quality to that generated by NTP.

\section{Conclusion}
\label{sec:conclusion}

In this paper, we introduced Diagonal Decoding~(DiagD), a training-free algorithm that significantly accelerates the inference speed of autoregressive video generation models. By leveraging spatial and temporal correlations in consecutive frames, DiagD generates tokens along diagonal paths, achieving substantial speedups while preserving visual fidelity. Through extensive experiments across diverse models, tasks, and datasets, we demonstrated the efficiency and generality of our approach, reducing the inference latency of Cosmos models by $10\times$ while maintaining their performance. Additionally, we proposed a lightweight fine-tuning strategy to close the training-inference gap, further improving generation quality with minimal computational cost. As a result, DiagD provides a practical and scalable solution for real-time video generation, pushing the boundaries of what is achievable with autoregressive Transformers in downstream tasks and related applications.

{
    \small
    \bibliographystyle{ieeenat_fullname}
    \bibliography{main}
}

\clearpage
\setcounter{page}{1}
\setcounter{figure}{0}
\setcounter{table}{0}
\maketitlesupplementary

\startcontents[chapters]
\setcounter{section}{0}
\printcontents[chapters]{}{1}{}

\section{Cosmos}
\subsection{Details of Cosmos Models}
Cosmos\citep{agarwal2025cosmos}, a World Foundation Model (WFM) Platform for developing Physical AI systems, integrates multiple pre-trained models, including autoregressive and diffusion-based methods, as well as discrete and continuous tokenizers. Specifically, the autoregressive model employs a discrete video tokenizer that leverages a codebook containing $16,000$ entries, achieving spatial compression of $16\times$ and temporal compression of $8\times$. This tokenizer is capable of compressing a video of $33$ frames at a resolution of $640\times1024$ into $12,800$ discrete tokens. In our study, we implement the diagonal decoding algorithm with both spatial and temporal acceleration on Cosmos autoregressive world foundation models. We provided an initial sequence of $5,120$ tokens (equivalent to $9$ frames), optionally accompanied by text depending on the task, to let the model generate subsequent $7,680$ tokens which represent $24$ frames. 

Due to the lack of an open-sourced evaluation pipeline and dataset, we replicated a comparable setup based on details provided in the technical report of Cosmos. Specifically, we randomly sample $100$ videos from the RealEstate10K dataset~\citep{zhou2018stereo} as the test set. To quantitatively assess visual quality, we employed standard metrics, including Fréchet Video Distance (FVD)~\citep{unterthiner2018towards} and Peak Signal-to-Noise Ratio (PSNR)~\citep{hore2010image}. Furthermore, we conducted a human evaluation, detailed in~\Cref{subsec:analysis}, to compare visual quality and object movement between videos generated by next-token prediction and DiagD. 

\subsection{Ablation Study on Hyper-Parameters}
We report results of DiagD with different combinations of $k$ and $d$ in~\Cref{sup:cosmos}. The findings show that $k$ has a more direct impact on performance than $d$, suggesting that excessive compression of spatial redundancy can lead to performance degradation due to accumulated errors. In contrast, temporal redundancy offers greater potential for exploration and utilization.

\begin{table*}[tb]
\centering
    \caption{Quantitative evaluation on Cosmos. We use $12$B models for video continuation tasks. "NTP" refers to the next-token prediction paradigm. $d=m, k=n$ denotes Diagonal Decoding algorithm with different hyper-parameter settings. "Step" refers to the number of forward passes required by the model to generate a video. "TP" is the number of tokens that model can generate per second on A$100$. }
    \label{sup:cosmos}
    % \resizebox{\columnwidth}{!}
        {
    	\begin{tabular}{l|cccc|ccc}
    		\toprule[1pt]
    	    Algorithm  & FVD$\downarrow$ & LPIPS$\downarrow$ & SSIM$\uparrow$ & PSNR$\uparrow$  & FPS$\uparrow$  &TP$\uparrow$ & STEP~($k$)$\downarrow$\\
            \midrule[1pt]
            NTP          & \bm{$135$} & 0.44 & 0.63 & 15.54  &0.15   &49   &$7.68$  \\
            \midrule
    		$d=1,k=1$    & 139 & 0.44 & 0.64 & 15.84  & \bm{$1.71$}  &\bm{$549$} &\bm{$0.11$}   \\
            $d=5,k=1$                 & 143 & 0.44 & 0.64 & 15.74  & \bm{$1.71$}       &\bm{$549$} &\bm{$0.11$} \\
            $d=9,k=1$                   & 136 & 0.44 & 0.63 & 15.68  & 1.61       &515    &0.12  \\
            $d=40, k=1$      & 136 & 0.44  & 0.63 & 15.71  &1.62   &512  &0.18 \\
            $d=2,k=2$              & 152 & \bm{$0.43$} & 0.64 & \bm{$15.98$} & 1.60  &512 &0.15 \\
            $d=10,k=2$               & 143 & 0.44 & 0.62 & 15.49 & 1.60  &512 &0.16 \\
            $d=18,k=2$             & 136 & 0.44 & 0.63 & 15.57   &1.50   & 480  & 0.18  \\
            $d=80, k=2$            & 136 & 0.44 & 0.63  & 15.59  &1.21   &384  &$0.30$  \\
            $d=4,k=4$               & 150 & \bm{$0.43$} & 0.64 & 15.86  &1.26   &427 &0.24 \\
            $d=12,k=4$              & 140 & 0.44 & 0.63 & 15.64  & 1.26  &404 &0.24 \\
            $d=20,k=4$                & 137 & 0.44 & 0.63 & 15.57  & 1.20  &384 &0.26 \\
            $d=36,k=4$            & 136 & 0.44 & 0.63 & 15.54 & 1.09  &349  &0.29 \\
            
            \bottomrule[1pt]
    	\end{tabular}
    }
    \label{tab:cosmos-result}
 \vspace{-3mm}
\end{table*}

\section{WHAM}
\subsection{Details of WHAM}
The World and Human Action Model (WHAM)~\citep{kanervisto2025world} is a recently proposed state-of-the-art autoregressive generative model trained on game data from \textit{Bleeding Edge}. The model is capable of generating coherent and diverse gameplay sequences based on user instructions. Unlike Cosmos, WHAM employs an image-level Vector Quantized (VQ) tokenizer that concentrates exclusively on spatial compression. This tokenizer independently converts each game state, with a resolution of $180\times 300$, into $540$ discrete tokens, which are subsequently concatenated with their corresponding in-game actions. To preserve the inherent relationship between actions and game states, we employ diagonal decoding with spatial acceleration alone. That is to say, we sequentially generate subsequent game states from previous states and their associated actions, alternating between state generation and action concatenation.

For WHAM, we randomly selected $100$ videos from its evaluation set to assess video consistency according to WHAM’s evaluation protocol. In each sample, the model is conditioned on one second of gameplay, which included previous video and controller actions, as well as the future actions performed by a human player, to generate consequent game states that follow these actions. 

\section{MC-AR}
\subsection{Details of MC-AR Models}
We conducted a series of experiments by training models from scratch on the VPT dataset~\citep{baker2022video}. The VPT dataset is a domain-specific dataset comprising gameplay videos from \textit{Minecraft}. We employed a pre-trained image VQ-VAE~\citep{patil2024amused}, an image-level tokenizer with a codebook containing $8,192$ entries, achieving a spatial compression ratio of $16\times$. To enhance visual quality, we subsequently fine-tuned the VQ-VAE on the VPT dataset. Our Transformer model was based on the LLaMA architecture~\citep{touvron2023llama} and augmented with 3D Rotary Embeddings~\citep{su2024roformer}. We combine each game state tokens with the corresponding actions just like WHAM, so for each pair of game state and actions in the original input $(x_i, a_i)$, the tokenizers will transfer them into a flat sequence of discrete ids as:
\begin{equation}
    (t_{i*c+1}, \cdots, t_{(i+1)*c}, t^{a_i}_1, \cdots, t^{a_i}_n).
\end{equation}
where $c$ is the number of ids to represent each state, $n$ is the number of actions.
We trained our model with next token prediction, enabling the model to predict future states based on previous game states and current action. We use the Adam optimizer\citep{kingma2014adam} with a cosine decay learning rate scheduler to train the model. Additionally, fine-tuning was performed for an extra $1,000$ steps on the same dataset.

For MC-AR, we selected $100$ video clips from an unused subset of the evaluation set, each containing $16$ frames, with the last $15$ frames corresponding to actions in the gameplay. Each model generated $15$ subsequent frames conditioned on the first frame of each clip and the $15$ actions. These generated frames were then compared against the ground truth using the FVD, PSNR, SSIM, and LPIPS metrics.

\subsection{Model Configurations}
\label{app:model_config}

\begin{table}[htbp]
    \centering
    \caption{The configuration of different size of models.}
    \resizebox{\columnwidth}{!}
    {
        \begin{tabular}{l|c|c|c|c}
            \toprule[1pt]
                & \textbf{Hidden dim} & \textbf{MLP dim} & \textbf{Num. Heads} & \textbf{Num. Layers} \\
            \midrule
            300M &  1024 & 4096 & 16 & 20  \\
            700M &  2048 & 4096 & 32 & 20 \\
            1.2B &  2048 & 8192 & 32 & 20 \\
            \bottomrule[1pt]
        \end{tabular}
    }
    \label{tab:model_arch}
\end{table}

\begin{table}[htbp]
    \centering
    \caption{Optimization hyperparameters.}
    \begin{tabular}{c|c}
        \toprule[1pt]
       \textbf{Hyperparameter}  &  \textbf{Value} \\
       \midrule
       Learning rate scheduler   &  cosine \\
       Learning rate & $3e^{-4}$ \\
       Warm up steps & 10000 \\
       Weight decay & 0.1 \\
       Optimizer & AdamW \\
       AdamW betas & $(0.9, 0.95)$ \\
       Maximum Positions & $5376$ \\
       \bottomrule[1pt]
    \end{tabular}
    \label{tab:hyperparams}
\end{table}

We train three different sizes of the model within the LLaMA architecture: $300$M, $700$M, and $1.2$B. We tune the hidden dimension, intermediate dimension, and the number of layers to achieve different model sizes. The configuration of these models are listed in Table~\ref{tab:model_arch}. The hyperparameters of the optimizer used to train the model are listed in Table~\ref{tab:hyperparams}.

\begin{table}[t]
    \caption{Quantitative evaluation on 300M MC-AR. We use DiagD~($k=1,2$) in experiment.}
    \label{tab:300minecraft-result}
    \centering
    \vspace{-2mm}
    \resizebox{\columnwidth}{!}
    {
    	\begin{tabular}{l|cc|cc}
    		\toprule[1pt]
    	    Algorithm  & PSNR$\uparrow$ & FVD$\downarrow$ & FPS$\uparrow$ & STEP~($k$)$\downarrow$ \\
    		\midrule
    		NTP                               &\bm{$15.63$}       &\bm{$223$}   &1.08    &5.04    \\
            DiagD\ $k=2$ w/o FT                &15.13               &246   &3.98    &0.75       \\
            DiagD\ $k=1$ w/o FT                &14.52               &500   &\bm{$4.29$}    &\bm{$0.56$}       \\
            DiagD\ $k=2$ w/ FT                 &15.22               &233   &3.98    &0.75     \\
    		\bottomrule[1pt]
    	\end{tabular}
    }
 \vspace{-2mm}
\end{table}

\begin{table}[t]
    \caption{Quantitative evaluation on 1.2B MC-AR. We use DiagD~($k=1,2,4$) in experiment.}
    \label{tab:1.2minecraft-result}
    \centering
    \vspace{-2mm}
    \resizebox{\columnwidth}{!}
    {
    	\begin{tabular}{l|cc|cc}
    		\toprule[1pt]
    	    Algorithm  & PSNR$\uparrow$ & FVD$\downarrow$ & FPS$\uparrow$ & STEP~($k$)$\downarrow$ \\
    		\midrule
    		NTP                               &\bm{$16.06$}       &\bm{$203$}   & 0.89   &5.04    \\
            DiagD\ $k=4$ w/o FT                &15.13               &246   &1.42        &1.14        \\
            DiagD\ $k=2$ w/o FT                &15.13               &246   &1.98       &0.75       \\
            DiagD\ $k=1$ w/o FT                &14.52               &500   &\bm{$2.48$}    &\bm{$0.56$}       \\
            DiagD\ $k=2$ w/ FT                 &15.30                &227   &1.98     &0.75     \\
    		\bottomrule[1pt]
    	\end{tabular}
    }
\end{table}

\subsection{Extra Results}
We provide results of models with different scales in Table~\ref{tab:300minecraft-result} and Table~\ref{tab:1.2minecraft-result}. The proposed Diagonal Decoding achieves consistent performance across scales, and fine-tuning always brings benefits.

\section{Derivations}
We derive Equation~(\ref{equ:ratio_spatial}) and (\ref{equ:ratio_diag}) here. First, for Equation~(\ref{equ:ratio_spatial}), assume $\min\{h,w\}=h$, we have:
\begin{equation}
\begin{split}
    r_{\textrm{spa}} &= \frac{h\cdot w}{(h-1)\cdot k + w} \\
    &= \frac{h}{\frac{h}{w}\cdot k - \frac{k}{w} + 1} \\
    &\approx \frac{h}{\frac{h}{w}\cdot k+1}.
\end{split}
\label{equ:ratio_spatial_derive}
\end{equation}
Where we assume $w \gg k$ and $\frac{k}{w}\approx 0$ which is applicable for most of our cases. 

Similarly, for Equation~(\ref{equ:ratio_diag}), we have:
\begin{equation}
\begin{split}
    r_{\textrm{diag}} &= \frac{T\cdot h \cdot w}{(T-1)\cdot h + h+w-1} \\
    &= \frac{T\cdot h \cdot w}{T\cdot h +w-1} \\
    &= \frac{w}{1+\frac{w-1}{T\cdot h}} \\
    &\approx w
\end{split}
\label{equ:ratio_diag_derive}
\end{equation}
Where the approximation stands when $T*h \gg w-1$, which is applicable for most of video generation cases.

\section{More Cases}
For more cases, please refer to our project page \url{aka.ms/diagd}.

\end{document}